\documentclass[10pt,twocolumn,letterpaper]{article}

\usepackage{cvpr}              

\usepackage{graphicx}
\usepackage{amsmath}
\usepackage{amssymb}
\usepackage{booktabs}
\usepackage{nidanfloat}     
\usepackage{makecell}       

\usepackage{algorithm}      
\usepackage{algpseudocode}  

\usepackage[pagebackref,breaklinks,colorlinks]{hyperref}

\usepackage[capitalize]{cleveref}
\crefname{section}{Sec.}{Secs.}
\Crefname{section}{Section}{Sections}
\Crefname{table}{Table}{Tables}
\crefname{table}{Tab.}{Tabs.}

\def\cvprPaperID{6859} 
\def\confName{CVPR}
\def\confYear{2022}

\begin{document}

\title{How to Guide Adaptive Depth Sampling?}

\author{Ilya Tcenov, Guy Gilboa\\
Technion - Israel Institute of Technology\\
{\tt\small \{ilya.tcenov, guy.gilboa\}@ee.technion.ac.il}
}
\maketitle

\begin{abstract}
Recent advances in depth sensing technologies allow fast electronic maneuvering of the laser beam, as opposed to fixed mechanical rotations. This will enable future sensors, in principle, to vary in real-time the sampling pattern. We examine here the abstract problem of whether adapting the sampling pattern for a given frame can reduce the reconstruction error or allow a sparser pattern. We propose a constructive generic method to guide adaptive depth sampling algorithms.

Given a sampling budget $B$, a depth predictor $P$ and a desired quality measure $M$, we propose an Importance Map that highlights important sampling locations. This map is defined for a given frame as the per-pixel expected value of $M$ produced by the predictor $P$, given a pattern of $B$ random samples. This map can be well estimated in a training phase. We show that a neural network can learn to produce a highly faithful Importance Map, given an RGB image.
We then suggest an algorithm to produce a sampling pattern for the scene, which is denser in regions that are harder to reconstruct. The sampling strategy of our modular framework can be adjusted according to hardware limitations, type of depth predictor, and any custom reconstruction error measure that should be minimized. We validate through simulations that our approach outperforms grid and random sampling patterns as well as recent state-of-the-art adaptive algorithms. 

\begin{figure*}[hb]
\centering
\includegraphics[width=1.0\textwidth]{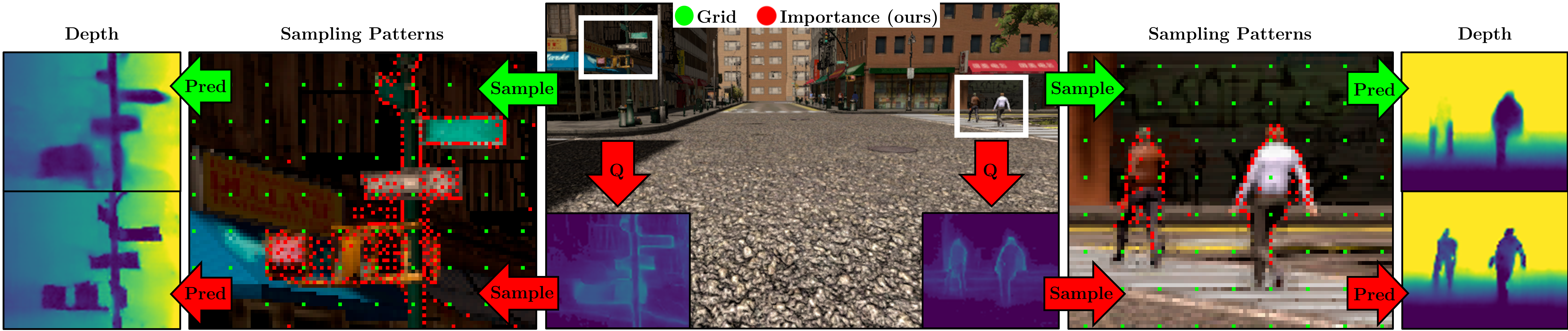}
\caption[Motivation]{Our adaptive depth sampling method guides the sampling process to obtain high frequency information which is crucial for accurate depth prediction. This results in sharper edges of the reconstructed objects and substantially lower error. The main component of our method is the $Q$ map which indicates a per-pixel importance for sampling for a given depth sensing system configuration.}
\label{fig:motivation}
\end{figure*}

\end{abstract}

\section{Introduction}
\label{sec:intro}

Estimating data based on limited amount of samples is a common problem in many scientific fields.
Gathering data is often expensive (in terms of power or time), especially when it involves active systems, such as computed tomography, radars and LiDARs. Adaptive sampling techniques have been proposed for Monte Carlo methods \cite{bucher1988adaptive}, sensor networks \cite{jain2004adaptive}, weather forecast \cite{bishop2001adaptive} and more. In the field of signal and image processing, compressed sensing \cite{donoho2006compressed,candes2006stable,eldar2012compressed} has been thoroughly developed, based on the notion of sparsity of the signal in a certain domain. Here we present a general sampling strategy for depth sensing, well adapted for deep learning methods. It does not assume any prior model on the depth scene and is based solely on the predicted error of a given depth completion algorithm.

\noindent {\bf Depth Completion.}
Conventional depth sensors
produce depth samples of the scene which are sparsely located. Accurately completing this sparse information to form a dense map is essential for many autonomous systems. This process is called \emph{depth completion} or \emph{depth reconstruction}. 
Initially, some classical approaches \cite{drozdov2016robust}, \cite{ma2016sparse}, \cite{ku2018defense} were proposed to solve the depth completion problem. In recent years, learning-based algorithms were proposed for the task. Finding a better CNN architecture for depth completion was the main focus of works such as \cite{ma2018sparse}, \cite{chen2018estimating}, \cite{cheng2018depth} and \cite{shivakumar2019dfusenet}. Other studies were aimed at proposing designated modules for solving the task \cite{tang2020learning,lee2020deep,lee2020deep,chen2019learning}.
The addition of an intermediate task is often used as a method for guidance to the learning process.
Lee \etal \cite{lee2019depth}, Xu \etal \cite{xu2019depth}, and Qiu \etal \cite{qiu2019deeplidar} compute surface normals as intermediate task. 
FusionNet by Van Gansbeke \etal \cite{van2019sparse} and the network of Eldesokey \etal \cite{eldesokey2019confidence} rely on guidance maps to increase the accuracy of the generated dense depth map. FusionNet has shown state-of-the-art results on the KITTI depth completion benchmark \cite{Uhrig2017THREEDV}. This network is used to showcase our approach as a prototypical modern depth completion algorithm. 

\noindent {\bf Adaptive Depth Sampling}
is a concept that can be applied to various fields. Badarna and Shimshoni \cite{badarna2019selective} developed selective sampling algorithms for decision trees and random forests. Dovrat \etal \cite{dovrat2019learning}, and Lang \etal \cite{lang2020samplenet} aim to pick a subset of points from a given point cloud in order to reduce computational complexity of a task, such as classification or registration.
Adjusting the sampling pattern according to the scene can be highly beneficial for depth reconstruction accuracy. Nevertheless, only few works proposed adaptive methods as a substitute for random, grid, or other fixed sampling strategies. Hawe \etal \cite{hawe2011dense} and Liu \etal \cite{liu2015depth} proposed sampling at locations that have high probability of high magnitude depth gradient. Wolff \etal \cite{wolff2019image} proposed a method that adopts the piece-wise planar model for depth representation. This approach shows impressive results, but does not utilize the advantages of deep neural networks.
Bergman \etal \cite{bergman2020deep} proposed a fully differentiable architecture for adaptive depth sampling that can be trained end-to-end. They target the extreme low-budget regime (and show that at sparsity above $\approx 0.1\%$  their advantage over fixed sampling diminishes).
Gofer \etal \cite{gofer2021adaptive} use the variance (disagreement) between an ensemble of predictors (neural networks or random forests) to extract the most relevant sampling locations. 
They focus on minimizing RMSE (root mean square error), which is closely related to variance. A strong feature in \cite{gofer2021adaptive} is their ability to work also without guidance. Our work, however, aims to go beyond RMSE, which may not be the most suitable measure to assess depth reconstruction. 

\noindent {\bf Image to Image Translation}
 transforms an image from one domain to another, while preserving the global (geometric) consistency. Usually, autoencoders, such as U-Net \cite{ronneberger2015u}, serve as a basis for the generator architecture design. The use of $L_1$ or $L_2$ losses often results in blurry images \cite{isola2017image}, \cite{johnson2016perceptual}. Thus, modern approaches are based on a discriminator CNN with adversarial loss \cite{goodfellow2014generative}, exhibiting higher fidelity translations. Conditional generative adversarial networks (cGANs) \cite{mirza2014conditional}, are commonly used in this context.
Image-to-image translation networks were used for various applications, such as image restoration \cite{luo2020time}, super resolution \cite{ledig2017photo}, inpainting \cite{pathak2016context}, and more \cite{dong2017semantic,zhang2017age,karacan2016learning,kaneko2017generative,sangkloy2017scribbler,zhu2017toward,wang2016generative}. Generation of high resolution images has shown to be a difficult problem. Inspired by \cite{chen2017photographic}, Wang \etal \cite{wang2018high} proposed Pix2PixHD. This multiscale architecture yields excellent translations of high resolution images for a wide range of modalities. 
We view the estimation of the proposed guidance measure ($Q$ map), based on an RGB frame,  as an image-to-image translation problem and use Pix2PixHD to solve it.

\noindent {\bf Paper Goal:} In our work we suggest a generic guidance measure for efficient adaptive depth sampling, which allows to optimize a wide variety of quality measures of a given depth completion method. Based on this guidance, a simple algorithm is proposed which compares favourably to \cite{wolff2019image} and \cite{gofer2021adaptive}. The measure can be further used by more sophisticated algorithms or as an additional loss-term in  network architectures. 

\section{Method}
\label{sec:method}

\subsection{Key Concept} 
A fundamental principle in data acquisition is that more information should be gathered where the uncertainty is larger. Intuitively, in a
polling example: areas in which people voted differently than expected  will require more extensive polling next time, compared to well-predicted regions.
Uncertainty can be measured by the expected value of the prediction error. This concept was used, for instance, in \cite{jain2004adaptive}
in the context of sensor networks.
In learning, a core idea of boosting techniques \cite{freund1999short} is similar, where more weight is given to data points which were miss-classified based on the current set of weak learners. For the depth sampling and reconstruction problem, this translates to the following principle:
\emph{Sample more densely in regions that are harder to reconstruct.}

Let us first formulate the adaptive sampling problem. 
For depth frame $y$, a sampling budget of $B$ samples is given. We assume pointwise sampling and that the sampling pattern can be selected arbitrarily. Additional constraints can be added, according to specific hardware limitations. Here we discuss the general unconstrained case. A sampling pattern $s$ measures $B$ points of $y$, denoted by $y(s)$. A predictor $P$ gets as input $y(s)$ and estimates the remaining unmeasured parts, yielding an estimated depth map $\hat{y}=P(y(s))$. For simplicity, we neglect measurement noise (it does not alter the essential principles of the model). Finally, an error measure  $M(a,b)$ defines a positive measure (generalized distance) between two depth frames $a$ and $b$. The adaptive sampling problem is to choose the best pattern $s^*$ for a given frame, such that the error is minimized, that is,
\vspace{0.2cm}
\begin{equation}
s^* = \underset{s}{\mathrm{argmin }} \{ M(y,P(y(s))) \}.
\label{eq:pattern_s}
\end{equation}
We restrict $M$ to be a function of a sum of a pointwise measure $q$ (this is valid for most error measures used in the context of depth completion). 
For example, for RMSE we have,
\begin{equation}
M_{RMSE} = \sqrt{\frac{1}{N}\sum_{i=1}^{N}\underbrace{(y_{i} - {\hat{y}_{i}})^{2}}_{\normalfont {q_{i}}}} \Longrightarrow q = (y_i - \hat{y_i})^{2},
\label{eq:M_rmse}
\end{equation}
and for REL (relative error),
\begin{equation}
M_{REL} = \frac{1}{N}\sum_{i=1}^{N} \underbrace{\frac{|{y}_i - {\hat{y}_{i}}|}{{y}_i}}_{\normalfont {q_{i}}} \Longrightarrow q = \dfrac{|y_i - \hat{y_i}|}{y_i}, 
\label{eq:M_rel}
\end{equation}
where $i$ is the index of a pixel ($N$ in total). 
Surely, problem \eqref{eq:pattern_s} cannot be easily solved. We have a ``chicken and egg'' problem. The optimal pattern is highly dependent on the unknown signal $y$. However, to obtain an estimation of $y$ we need to sample it. One may resort to iterative sampling, as suggested in \cite{gofer2021adaptive}. Here we propose to use RGB side information, in order to estimate locations of high uncertainty, which require denser sampling. 

We now address the issue of estimating the uncertainty.
A depth completion neural network entails very complex modeling of the depth signal, along with nuanced correlations to the RGB image (for guided algorithms).
Thus, any notions of general expected uncertainty, such as proximity to depth edges, would only amount to a first order gross approximation. Specifically, the extent of uncertainty in a region can best be evaluated by the depth completion algorithm itself. We thus propose to measure the expected uncertainty using the predictor $P$, given a neutral sampling pattern. We use uniform random sampling for that. 
Let $s_r$ denote a uniform random sampling pattern of $B$ samples.
We define $Q^*$ of a depth frame $y$ as the expected value of $q$ given a random pattern $s_r$,
\begin{equation}
Q^*(y) = Q^*(y\,|\,P,B,q):=E[q\,|\,s_r],
\label{eq:Q_star}
\end{equation}
where $E$ denotes the pointwise expected value over all $s_r$ patterns. This uncertainty map can certainly be estimated for a training set. 
Let us define a set of $J$ random sampling patterns $\{s_r^j\}_{j=1}^J$. We denote by $\hat{y}^j := P(s_r^j)$ the corresponding prediction for each sampling pattern and by $q^j$ the respective error.
Our estimation $Q \approx Q^*$ is the empirical expected value, performed by simple averaging,
\begin{equation}
Q := \frac{1}{J}\sum_{j=1}^{J}q^j.
\label{eq:Q}
\end{equation}

\begin{figure}
\begin{centering}
  \begin{subfigure}{0.49\columnwidth}
    \includegraphics[width=\textwidth]{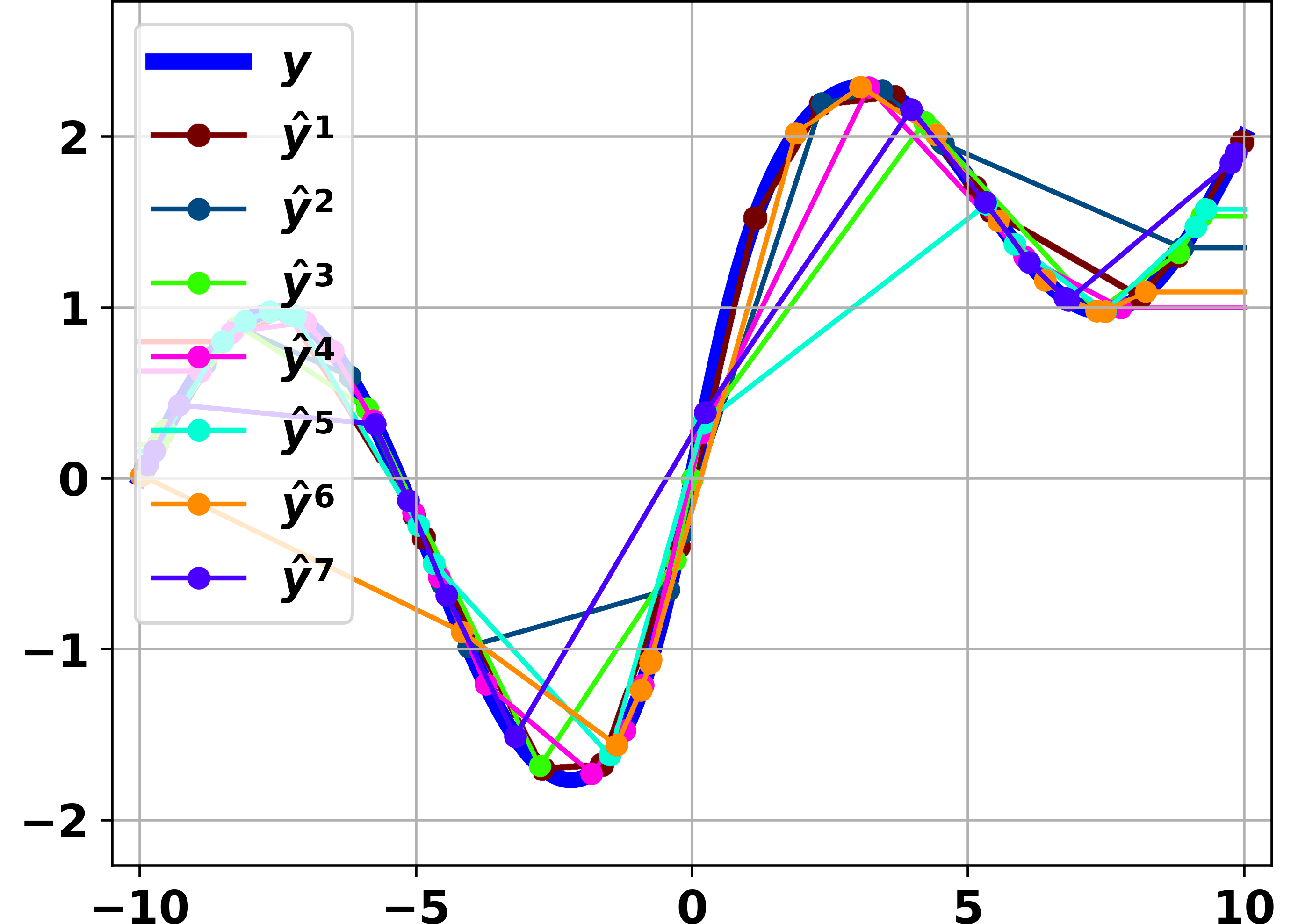}
    \caption{\footnotesize Non-adaptive sampling.}
    \label{fig:error_based_sampling:non_adaptive}
  \end{subfigure}
  \begin{subfigure}{0.49\columnwidth}
    \includegraphics[width=\textwidth]{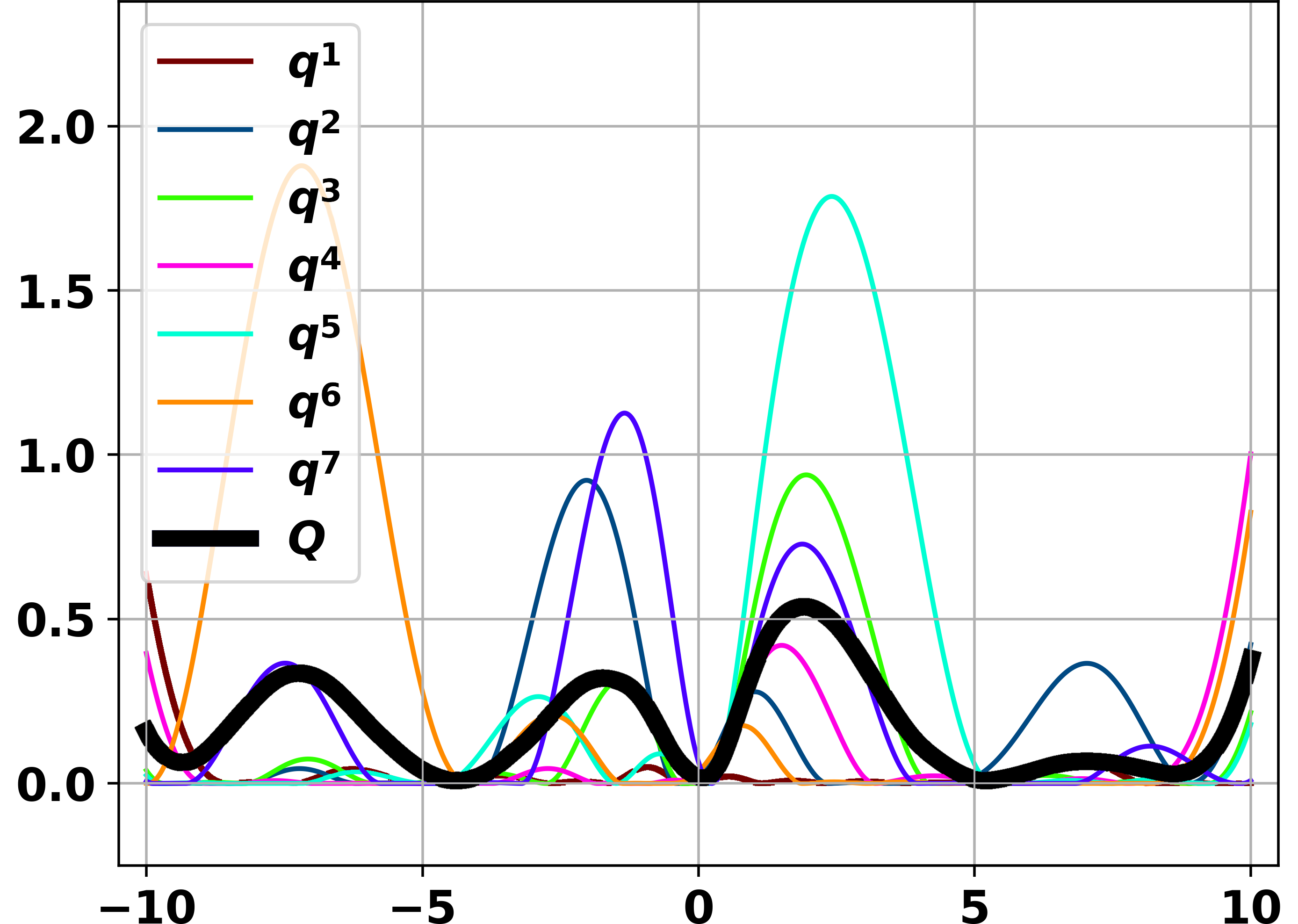}
    \caption{\footnotesize Reconstruction error.}
    \label{fig:error_based_sampling:q}
  \end{subfigure}
  \begin{subfigure}{0.49\columnwidth}
    \includegraphics[width=\textwidth]{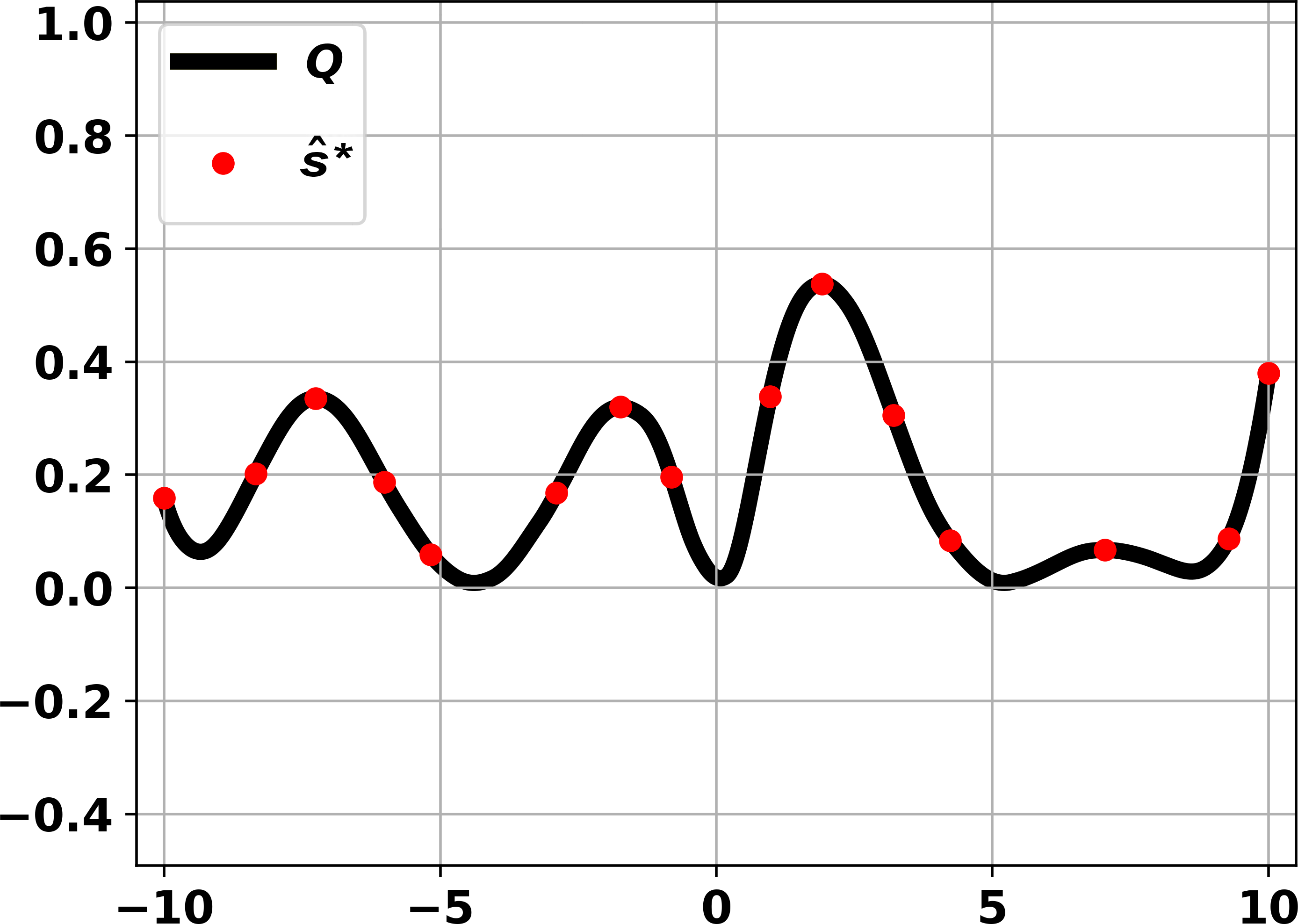}
    \caption{\footnotesize Adaptive pattern.}
    \label{fig:error_based_sampling:Q}
  \end{subfigure}
  \begin{subfigure}{0.49\columnwidth}
    \includegraphics[width=\textwidth]{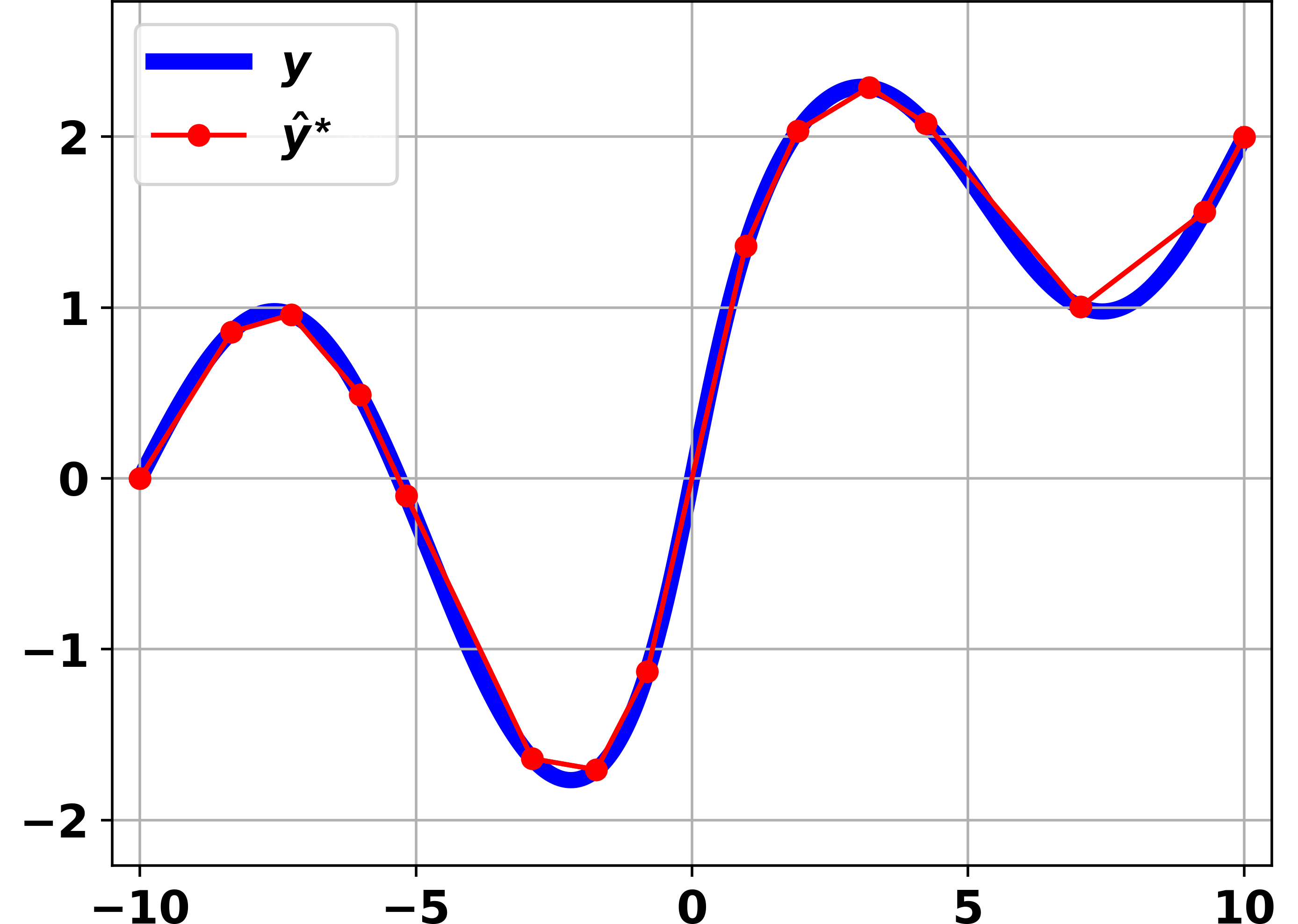}
    \caption{\footnotesize Adaptive sampling.}
    \label{fig:error_based_sampling:adaptive}
  \end{subfigure}
\caption[Error-based adaptive sampling toy example]{\textbf{(a)}
The signal $y$ (blue) along with 7 reconstructions $\hat{y}^j$ (linear interpolation), based on different random sampling patterns (average RMSE = 0.392). \textbf{(b)} A pointwise error $q^j=(y-\hat{y}^j)^2$ for each $\hat{y}^j$ and the mean $Q$ (black). \textbf{(c)} High values of $Q$ are marked as important, with adequate spacing, yielding an adaptive sampling pattern $\hat{s}^*$ (red dots). \textbf{(d)} Reconstruction $\hat{y}^*$ (red) based on $\hat{s}^*$ 
and the same interpolator yields significantly lower error (RMSE = 0.075).\vspace{1.4cm}}
\label{fig:error_based_sampling}
\end{centering}
\end{figure}

Let us illustrate the relation between the expected error and the sampling pattern in a simple toy example,
shown in Fig. \ref{fig:error_based_sampling}.
In this example we are given a budget of $B=15$ samples and are required to choose the best sampling pattern to reconstruct a one-dimensional signal. Our interpolator $P$ is linear (Fig. \ref{fig:error_based_sampling:non_adaptive}). To illustrate the informative value of the prediction error we assume here to have an oracle which returns the pointwise square error $q^j$, as defined in Eq. \eqref{eq:M_rmse}, for a given interpolation $\hat{y}^j$ (Fig. \ref{fig:error_based_sampling:q}). The empirical expected error $Q$ can now be calculated (Fig. \ref{fig:error_based_sampling:Q}). In an actual application $Q$ is estimated (we use RGB for that). It is shown that $Q$ has high values near local minima and maxima of $y$, where linear interpolation yields large errors. Sampling at local maxima of $Q$, with some required spacing (solved in our proof-of-concept by Gaussian Sampling) yields a highly effective sampling pattern (Fig. \ref{fig:error_based_sampling:adaptive}), which reduces the prediction error by about a factor of 5.

\subsection{Estimating $Q$ from RGB}
We propose a method for estimating a $Q$ map, given an RGB image, registered to the depth frame. This can provide crucial information for the guidance of adaptive sampling algorithms.
It is possible to extend this idea to other modalities as well, such as estimation based on thermal sensors or radars.
In order to demonstrate how this information can be exploited, we propose in Sec. \ref{sec:poc_stages} an end-to-end proof-of-concept (the final inference system is shown in Fig. \ref{fig:method}).

The main challenge, naturally, is to estimate $Q$ for a given frame, as $Q$ can be computed only when a ground truth is available.
We treat $Q$ as an image of the scene of a different modality, casting the task as an image-to-image translation problem. We found Pix2PixHD by Wang \etal \cite{wang2018high} yields a highly faithful translation. The network takes an RGB image as input and outputs an estimation of $Q$.
We denote this network as $RGB2Q$.  

In order to compute $Q$ in a training set we need to fix the sampling budget $B$, to decide on the relevant error measure $M$ (and hence $q$) and to choose the predictor $P$, which is later used by the final algorithm to estimate a full depth map from the sparsely sampled measurements.
For $P$, the reconstruction network, we choose FusionNet  by Van Gansbeke \etal \cite{van2019sparse}, which is RGB-guided. Based on the inputs of sparse depth samples and a corresponding RGB image, it reconstructs the missing depth values, generating a dense depth map. 
\emph{Our goal} amounts to increasing the depth reconstruction accuracy (reducing the error) by providing the most relevant sampling pattern of $B$ samples, per frame, as shown in Fig. \ref{fig:motivation}.

{\bf Limitations of the proposed framework:} We note there are a few assumptions, related to the LiDAR data, which simplify the problem and was assumed by us and by recent papers in the field \cite{wolff2019image,bergman2020deep,gofer2021adaptive}. First, it is assumed that the adaptive LiDAR can sample depth at any specified location (for KITTI dataset the ground truth is not full and the assumption is weaker). In addition, the LiDAR is capable of pointwise sampling (the main lobe of the illumination is a delta function). The sampling noise is negligible. Moreover, there are no constraints on the sampling pattern which are dictated by a specific hardware configuration. These assumptions allow us to use the Synthia\footnote{Synthia Video Sequences: \url{http://synthia-dataset.net}} \cite{Ros_2016_CVPR} synthetic RGBd dataset.
Naturally, in real-world scenarios additional assumptions and constraints are required to guide a physical system. These constraints change according to the hardware specifications, cost, power limitations and the measurement technology.  This work treats the problem in a more abstract manner and investigates the benefits and costs of adaptive depth sampling, which is still a new research field.
Other limitations are the need to use simulated data, as in real-world benchmarks (such as KITTI) the GT depth is sparse and one cannot control well the chosen pattern. Our error measure $M$ is assumed to be based on a pixel-wise measure $q$, which is the common case. However, this does not include more complex semantic global measures.

\subsection{Framework Hyper-Parameters}
\label{sec:hyperparam}
In order to loosely incorporate general hardware and system constraints, our framework allows the tuning of 3 hyper-parameters (We put in parenthesis the values used in our experiments):
\begin{itemize}
    \item Sampling Budget $B$: number of depth samples per frame (9728). 
    \item Depth Threshold $DT$: the maximum depth value that can be measured (100m).
    \item Error Metric $M$:  depth reconstruction error measure to be minimized (RMSE / REL).
\end{itemize}
The first two ($B, DT$) are dictated by the hardware limitations of the sensor.
These dictate the speed at which the sensor can measure depth points, which can be translated to a per frame sampling budget $B$.
The third ($M$) is related to the task and tolerance of the algorithms using the data. 
We use a sampling percentage of $1\%$ (compared to the RGB resolution), which is low but still yields reconstructions of high quality and resolution (somewhat similar to \cite{wolff2019image,gofer2021adaptive}). We found that the extremely low-budget regime (below $0.1\%$, as in \cite{bergman2020deep}) is too coarse and loses significant information. This translates to a sampling budget $B$ of 9728 sampled pixels from the 1280x760 dense depth $GT$ at 5 FPS, or $48640$ samples per second. This sampling budget is significantly lower then what modern LiDAR technology is capable of. It potentially allows a drastic reduction in their size and cost.

Depth sensor accuracy deteriorates with distance. This accuracy is usually bounded to hardware limitations of the sensor. In order to take into account this real-world limitation in our framework (which is trained and tested on a synthetic dataset), we discard any sampled depth value above the defined depth threshold $DT$. We chose a realistic depth threshold ($DT = 100$ m).

The definition of the error metric $M$ that should be minimized is related to the task and requirements of the specific application. We chose to test our framework under two types of error metrics: $RMSE$ (Eq. \eqref{eq:M_rmse}) and $REL$ (Eq. \eqref{eq:M_rel}). Only valid pixels within the measuring range are considered.

\subsection{Algorithm Stages}
\label{sec:stages}

Our goal is to design a system capable of estimating a $Q$ map for a given RGB image. The algorithm steps are described in detail below.

\noindent {\bf Step 1 - Training a basic reconstruction network.}
We first train a reconstruction network,  FusionNet  \cite{van2019sparse},
where the sparse depth input consists of uniformly distributed random sampling patterns of budget size $B$.
We denote this network as $DepthReconRand$. 
The loss function of the network is set according to the predefined hyper-parameter $M$, as defined in Eqs. \eqref{eq:M_rmse} and  \eqref{eq:M_rel}. 
Once the network is trained, we obtain an image-guided depth reconstruction network $P_{rand}$, optimized for random sampling, where for a given  RGB image and a sampling pattern $s_r$ the depth estimation is,
\begin{equation}
\label{eq:y_rand}
    \hat{y} = P_{rand}(RGB, s_r).
\end{equation}
\noindent {\bf Step 2 - Calculating $Q$ maps.}
Following Step 1, we are able now to compute $Q$ maps as shown in Alg. \ref{alg:Qmap}. For a given pair of RGB image and ground truth depth map $y$, taken from the training data, a set of $J$ random sampling patterns is generated, $\{s_r^j\}_{j=1}^J$. The map $Q$ is computed by Eq. \eqref{eq:Q}, where $\hat{y}^j$ is computed by  \eqref{eq:y_rand} and $q$ is determined by the desired error measure, given in \eqref{eq:M_rmse} or \eqref{eq:M_rel}.
We have tested the impact of the size of the set $J$ on the computation of $Q$ and found that it converges around $J=100$ (the chosen value). Comparing the values of $Q$ for $J=1000$ and $J=10000$, very similar results were obtained. 
We note that in regions of distant objects (above $DT$) $Q$ is set to zero, so that zero importance is assigned for sampling far objects (beyond the range of the sampling device). 

\begin{algorithm}[htb] 
\caption{$Q$ Map Calculation}
\label{alg:Qmap}
\begin{algorithmic}[1]
\Require{$RGB$ image, ground truth depth image $y$, number of iterations $J$, $DT$, $B$,
depth reconstruction network $DepthReconRand$} 
\Ensure{$Q$ map}
\Statex
\Function{CalcQ}{Input}
    \State {Initialize $Q$}
    \For{$j$ = 1,2,\ldots,$J$}                    
        \State {$s_r^j$ $\gets$ randomly chosen $B$ samples of $y$}
        \State {$\hat{y}^j \gets DepthReconRand(RGB, s_r^j)$}
        \State {$Q \gets Q + q^j(y,\hat{y}^j)$ as in Eqs. \eqref{eq:M_rmse}, \eqref{eq:M_rel}}
    \EndFor
    \State {$Q[y > DT] \gets 0$}
    \State {$Q \gets Q / J$}
    \State \Return {$Q$}
\EndFunction
\end{algorithmic}
\end{algorithm}

\begin{figure}[htb]
\begin{centering}
  \begin{subfigure}{0.3\columnwidth}
    \includegraphics[width=\textwidth]{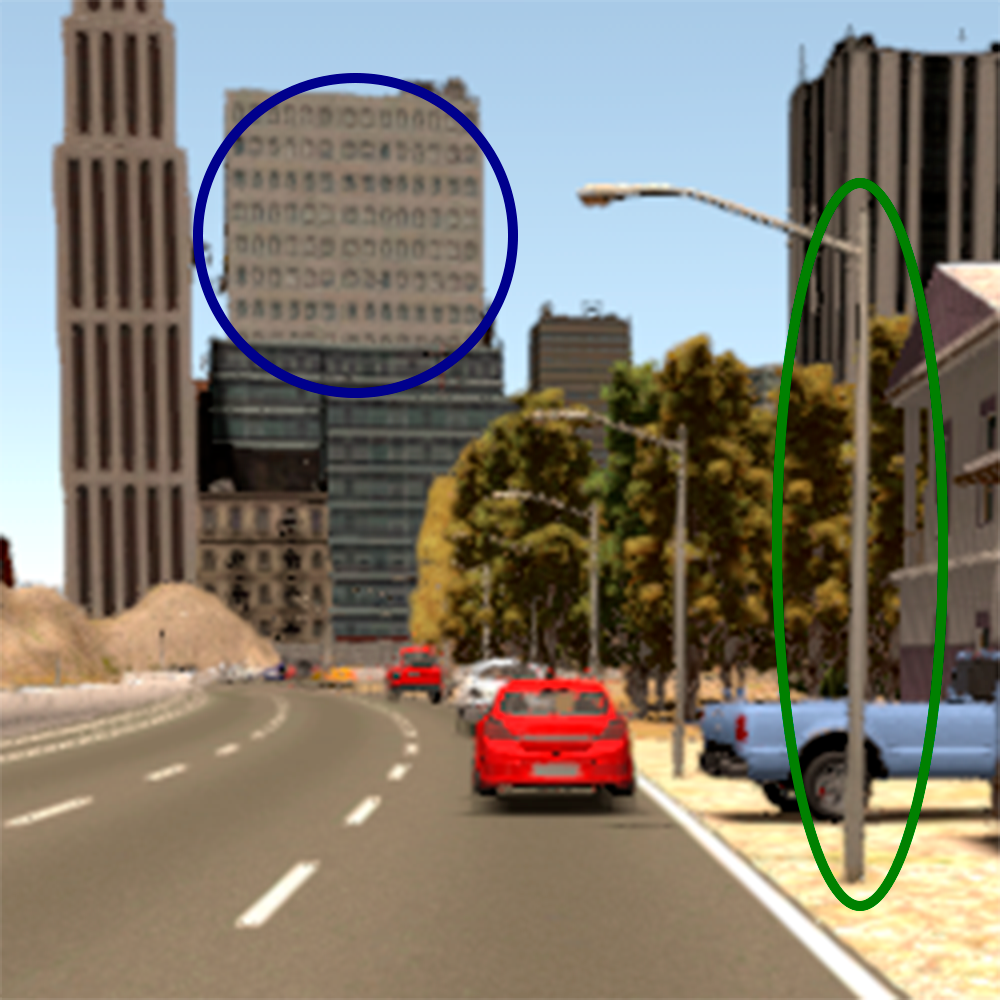}
    \caption{RGB image}
    \label{fig:rgb_example}
  \end{subfigure}
  \hspace{1cm}
  \begin{subfigure}{0.3\columnwidth}
    \includegraphics[width=\textwidth]{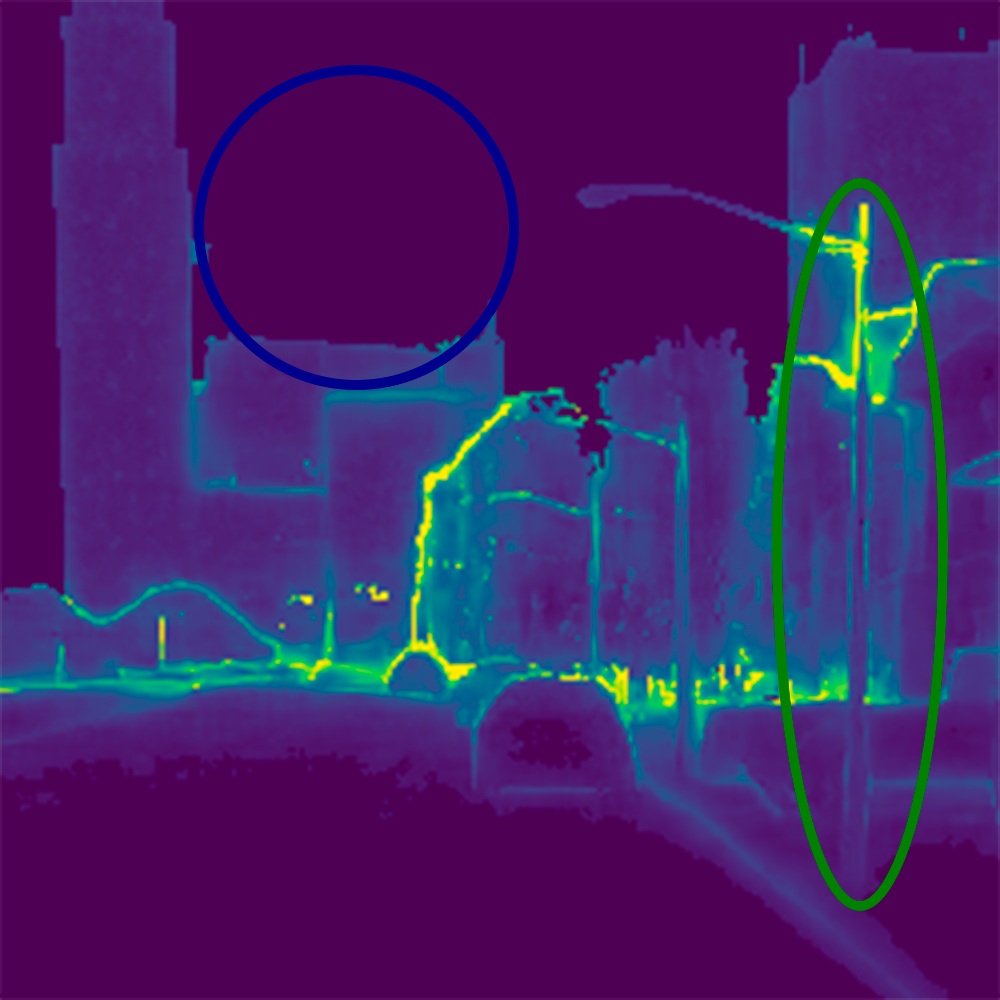}
    \caption{$\hat{Q}$ map}
    \label{fig:Q_example}
  \end{subfigure}
 \caption[Features of $\hat{Q}$]{An RGB patch and the corresponding generated $\hat{Q}$ are presented to demonstrate some of the features of $\hat{Q}$. Blue circle - 0 importance of a distant object. Green circle - varying importance that correspond to varying depth gradient magnitude.}
\label{fig:rgb-impmap_example}
\end{centering}
\vspace{-0.5cm}
\end{figure}

\noindent {\bf Step 3 - Estimating $Q$ from RGB.}
We have obtained pairs of $\{RGB,Q\}$ for every frame in the training data. This serves as a training set for the image-to-image translation network, Pix2PixHD \cite{wang2018high}.
More information regarding the training process is presented in the supplementary document.
At inference, the network provides $\hat{Q}$, an estimation of $Q$, given an RGB image. An example of such estimation is shown in \cref{fig:rgb-impmap_example}. The network is referred to as $RGB2Q$, as shown in Fig. \ref{fig:method}.

\begin{figure}[t]
\centering
\includegraphics[width=1.0\columnwidth]{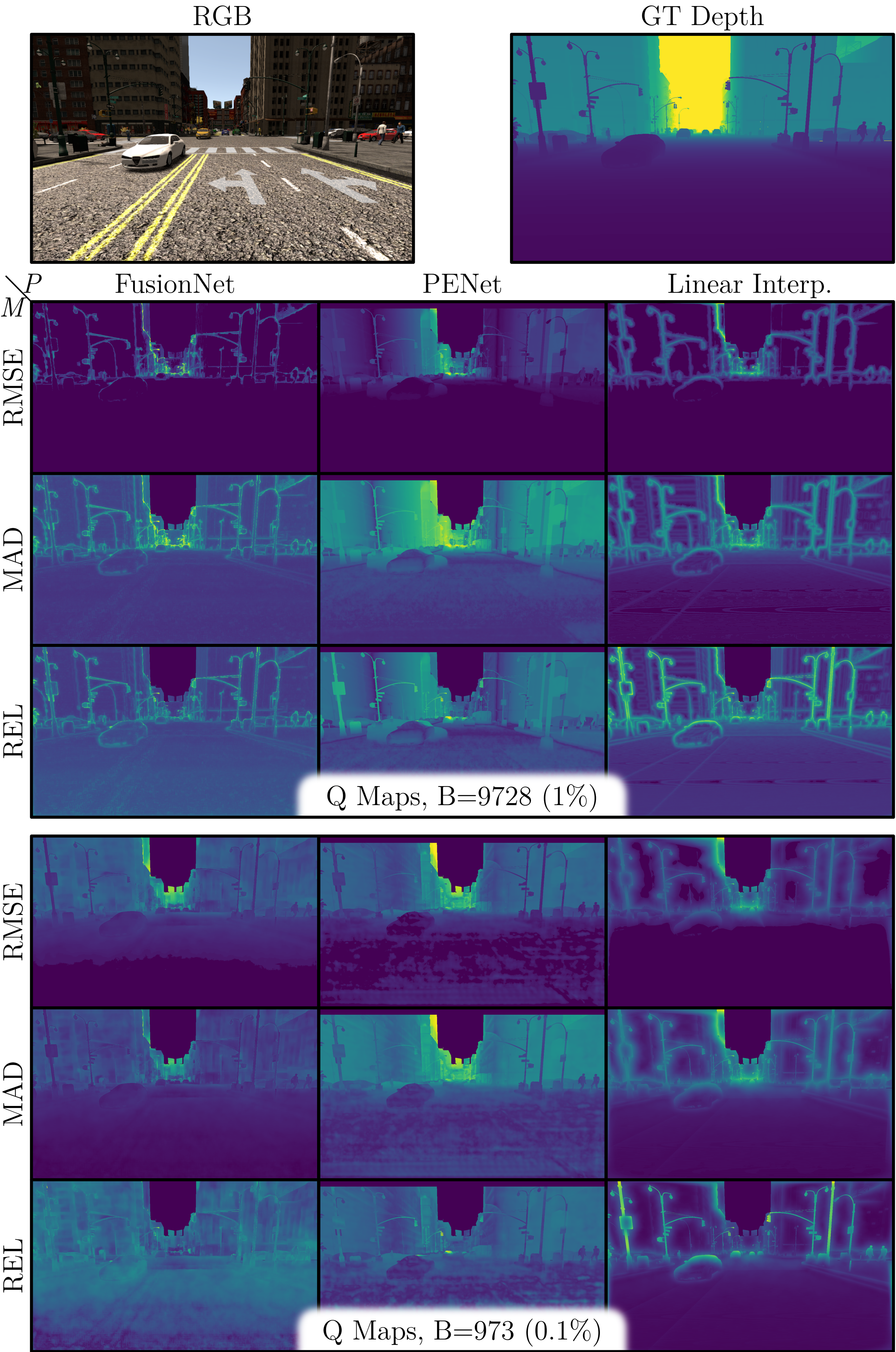}
\caption{A comparison of $Q$ maps for different $B$ (9728, 973), $P$ (FusionNet \cite{van2019sparse}, PENet \cite{hu2021penet}, linear interpolation) and $M$ (RMSE as in \cref{eq:M_rmse}, REL as in \cref{eq:M_rel}, MAD where $q$ is defined as a per-pixel absolute difference). The $Q$ maps were processed for comfortable viewing while preserving relative importance.}
\label{fig:impmaps_for_P_and_M}
\end{figure}

\paragraph{}
\noindent {\bf Remarks on the nature of $Q$.}
The robustness of the proposed approach lies in a broad space of predictors $P$ and error metrics $M$ that can be used to guide the depth sampler s.t. $M$ will decrease for the prediction of $P$. Almost any $P$ and $M$ can be used to produce corresponding $Q$ maps. 
Examples of $Q$ maps for different combinations of $B$, $P$ and $M$ are shown in \cref{fig:impmaps_for_P_and_M}. In addition to FusionNet, our default depth predictor $P$, PENet \cite{hu2021penet} was used as alternative neural network to create $Q$ maps for comparison. Similarly, linear interpolation was used as a simple algorithm for $P$. Brighter regions of $Q$ correspond to regions that statistically are harder to reconstruct. In other words, $Q$ map shows the weaknesses of $P$ under the error metric $M$.

As \cref{fig:impmaps_for_P_and_M} shows, linear interpolation struggles with reconstruction of depth gradients, since it does not exploit RGB spatial information and the generalization capabilities of neural networks. The $Q$ maps for FusionNet show a comparable behaviour, but these maps are affected by RGB elements since they are used in the prediction process (such as the yellow road marking). PENet on the other hand requires a less intuitive sampling pattern to increase the reconstruction accuracy. Distant objects are the most vulnerable for reconstruction inaccuracies. These reconstruction errors are most evident on $Q$ maps for $RMSE$ ($q = {error}^2$). Similar, but less noticeable effect is present on $MAD$ (mean absolute difference) $Q$ maps ($q = |error|$). A more homogeneous distribution is shown in $Q$ maps for $REL$, since the prediction error is normalized by depth value per-pixel.
As the sampling budget $B$ is decreased (by a factor of 10 in our example), the prediction accuracy of all methods is reduced in a global manner and the $Q$ maps are blurred as a result. Consequently, at such low sampling budgets the incentive to employ adaptive sampling strategies is reduced.

\begin{figure}[t]
\centering
\includegraphics[width=1.0\columnwidth]{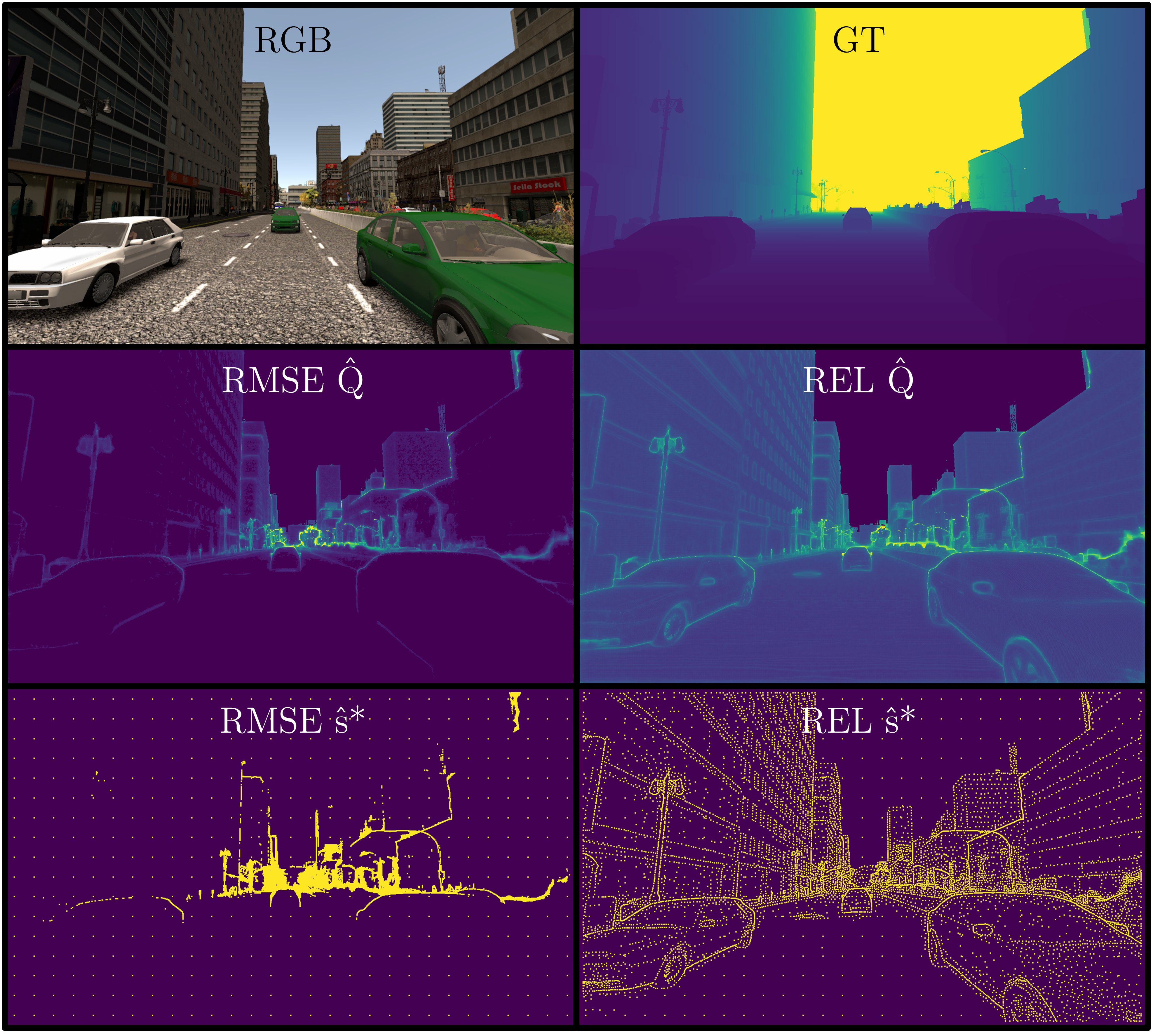}
\caption[A comparison of $ImpMaps$]{A comparison of $\hat{Q}$ and $\hat{s}^*$ for $M$ = RMSE and REL.}
\label{fig:impmapcomp}
\end{figure}

Different error measures $M$ lead to different characteristics of $Q$. For $M$ = RMSE, $Q$ maps have higher values in further regions of the scene. This is due to the fact that distant regions are harder to estimate accurately in terms of nominal error.  For $M$ = REL a more balanced $Q$ is obtained that results in a more uniform sampling pattern, as shown in Fig. \ref{fig:impmapcomp}. In both cases, depth edges are densely sampled according to the guided sampling algorithm which is proposed as a proof-of-concept. The RGB2Q model mimics this behaviour, therefore $\hat{Q}$ shows similar properties.


\section{Proof of Concept}
\label{sec:poc_stages}

\begin{figure*}[ht]
\includegraphics[width=1.0\textwidth]{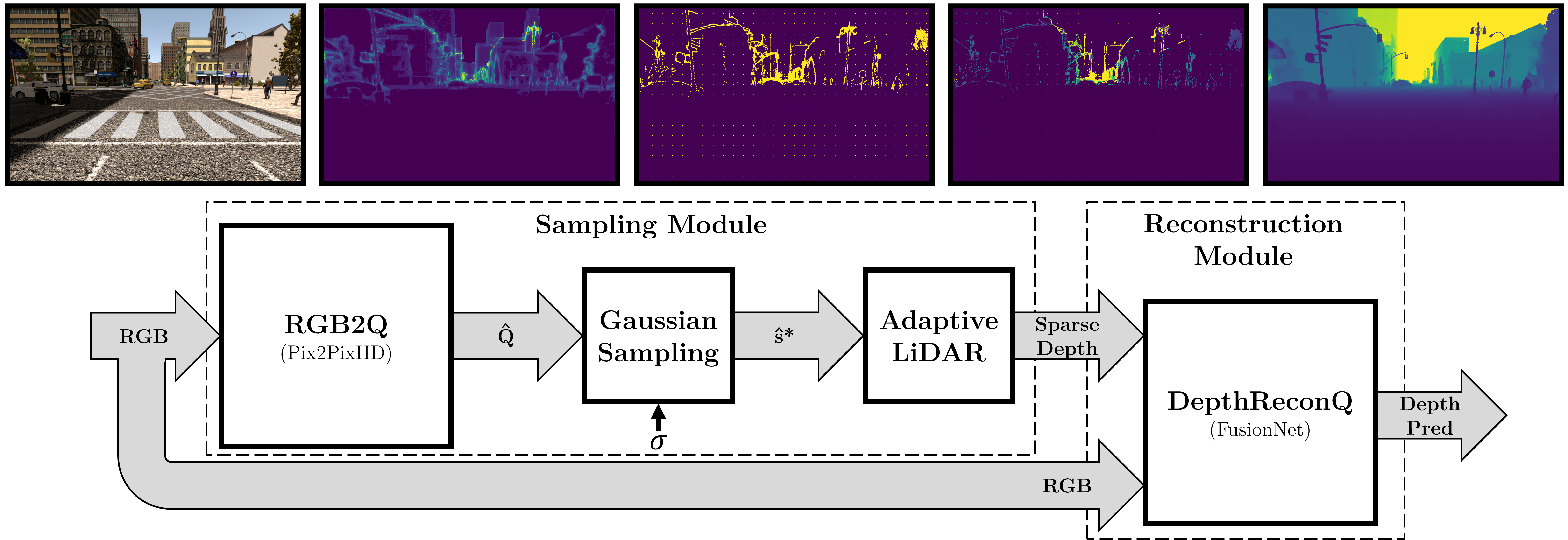}
\caption[Our framework]{Our inference framework. To demonstrate a method of utilization of the information stored in $\hat{Q}$, we propose an inference framework as a proof-of-concept. Gaussian Sampling is used to extract a sampling mask $\hat{s}^*$ which is then guide the depth sampling process s.t. the acquired information is crucial for the specific depth reconstruction algorithm.}
\label{fig:method}
\end{figure*}

Now that we have a good estimation of $Q$ maps we propose an adaptive sampling algorithm, guided by these maps.
For the sake of consistency, the enumeration of these steps is made as a continuation of the proposed method for $Q$ map generation.
These steps will allow us to design the full adaptive depth sampling inference framework (\cref{fig:method}).

\noindent {\bf Step 4 - Gaussian Sampling and coarse grid.}
Our aim is to obtain an adaptive sampling pattern
$\hat{s}^*$ based on $\hat{Q}$. 
We denote the process of generating the sampling mask as \emph{Gaussian Sampling}. The goal in this process is to select $B$ significant samples based on a given $\hat{Q}$ map. Selecting these samples in a straightforward manner as the $B$ maximal values of $\hat{Q}$ leads to clustered samples and suboptimal results. We should take into account the fact that when a point is sampled -- its close vicinity can usually be well inferred by the predictor $P$. Thus, our process adapts the map iteratively and attenuates regions close to a chosen sample. For simplicity, we take an iterative greedy approach. At each iteration the global maximum of $\hat{Q}$ is selected as a sampling point. The values of $\hat{Q}$ are then attenuated around this point by the complement of a Gaussian kernel, 
\begin{equation}
    k_\sigma(x,y) = 1 - e^{- \frac{x ^ 2 + y ^ 2}{2 \sigma ^ 2}}.
    \label{eq:gauss}
\end{equation}
For a sample chosen at $(x_j,y_j)$ the update of $\hat{Q}$ is,
\begin{equation}
    {\hat{Q}(x,y)}_{new} = {\hat{Q}(x,y)}_{old} \cdot
    k_\sigma(x - x_j,y - y_j).
    \label{eq:imp}
\end{equation}
Numerically, the kernel size is  $5\sigma \times 5\sigma$ pixels. 
Our results indicate that using the entire sampling budget $B$ based on $\hat{Q}$ results in sparse depth information that contains high frequency details, but lacks global information. In order to avoid situations in which entire scene regions remain unsampled, we use some portion of the budget $B$, denoted as $\alpha$, for a coarse grid sampling mask. We found that using 5\% of $B$ ($\alpha =0.05$) in grid sampling leads to better reconstruction.
Examples of result sampling patterns are presented in Fig. \ref{fig:impmapcomp}.
More information on Gaussian Sampling, including the method of finding the value of $\sigma$, can be found in the supplementary document.

\noindent {\bf Step 5 - Training the Final Reconstruction Network.}
After extracting highly relevant samples $\hat{s}^*$ in the previous step, we use them to train the depth reconstruction network $DepthReconQ$.
The training of this network is done similarly to $DepthReconRand$ (step 1). The only difference is that $DepthReconQ$ is trained based on the adaptive sampling pattern described in Step 4. All patterns contain $B$ samples each. The full system is depicted in \cref{fig:method}.

\section{Evaluation}
\label{sec:evaluation}

The algorithm steps can be reproduced using our code\footnote{GitHub repository: {\scriptsize \url{https://github.com/tWXWbAdq/How-to-Guide-Adaptive-Depth-Sampling.git}}}. We now evaluate the sampling method described in the previous section.
We refer to the sampling pattern $\hat{s}^*$ as "Importance pattern", since it reflects points which are important for sampling. The sampling pattern $\hat{s}^*$ guides an adaptive LiDAR that measures depth values at the requested sampling locations.
As mentioned before, we assume that LiDAR noise is negligible. This sparse depth map, along with an RGB image, is passed to the reconstruction network $DepthReconQ$ that predicts a dense depth map.

In order to compare the results of our framework that is based on image-guided sampling patterns to other sampling patterns (random, grid, SuperPixels\footnote{Center of mass of SLIC SuperPixels as shown by Wolff \etal \cite{wolff2019image}}), we train two additional reconstruction networks - $DepthReconGrid$ and $DepthReconSP$. These networks are not part of the framework, but have to be trained for a fair comparison of the reconstruction results. The training of these networks is done similarly to the training of $DeapthReconRand$ (step 1 in Sec. \ref{sec:stages}), while the corresponding sampling patterns are used in this process.

Each sampling pattern (random, grid, superpixels, importance) contains $B$ samples, so that we compare patterns with an equal number of samples. The random sampling pattern was made by randomly (uniform distribution) selecting $B$ samples. The grid mask was formed in a square pattern so that the samples are evenly spaced. The SuperPixel image-guided pattern \cite{wolff2019image} was composed by dividing the RGB image into $B$ SLIC super-pixels, and sampling at the center of mass of each super-pixel. 

\begin{figure}
\begin{centering}
  \begin{subfigure}{1.0\columnwidth}
    \includegraphics[width=0.9\columnwidth]{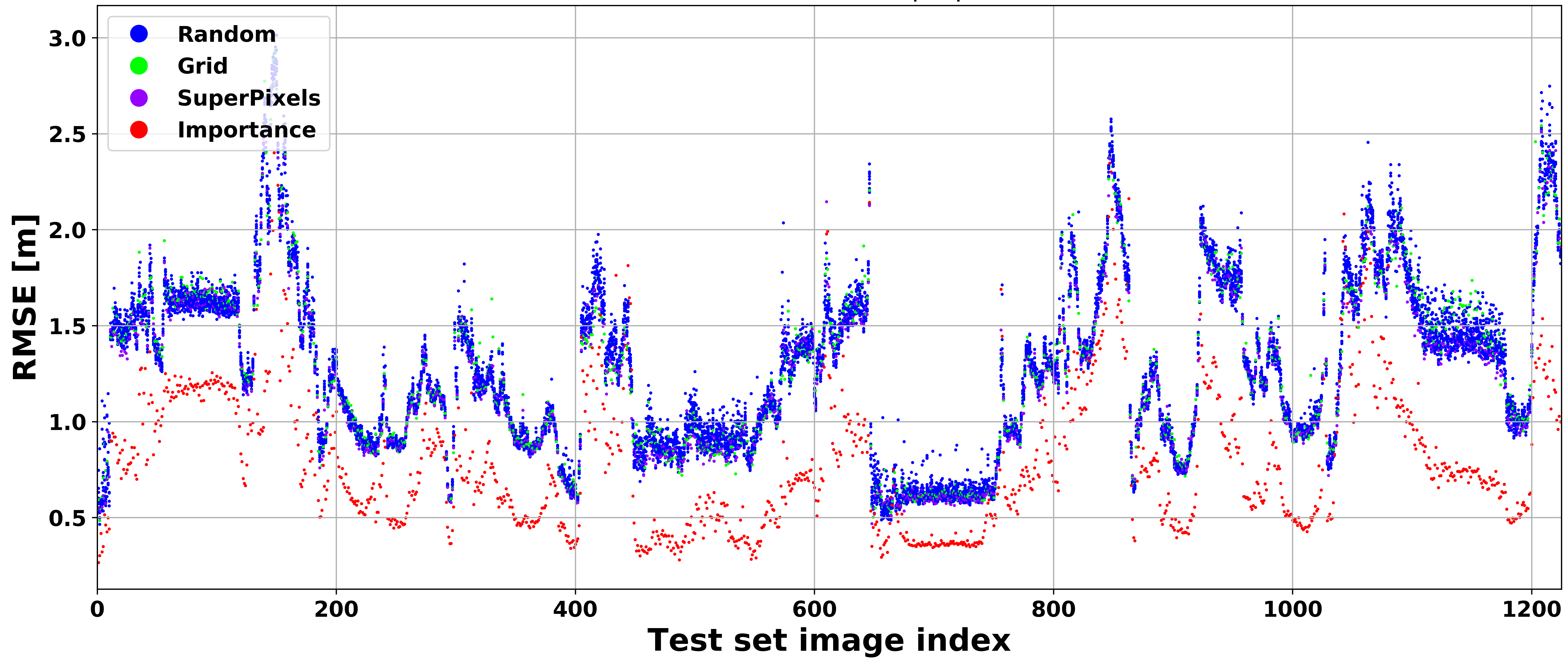}
    \caption[Reconstruction results - RMSE]{RMSE}
    \label{fig:res_rmse}
  \end{subfigure}
  \begin{subfigure}{1.0\columnwidth}
    \includegraphics[width=0.9\columnwidth]{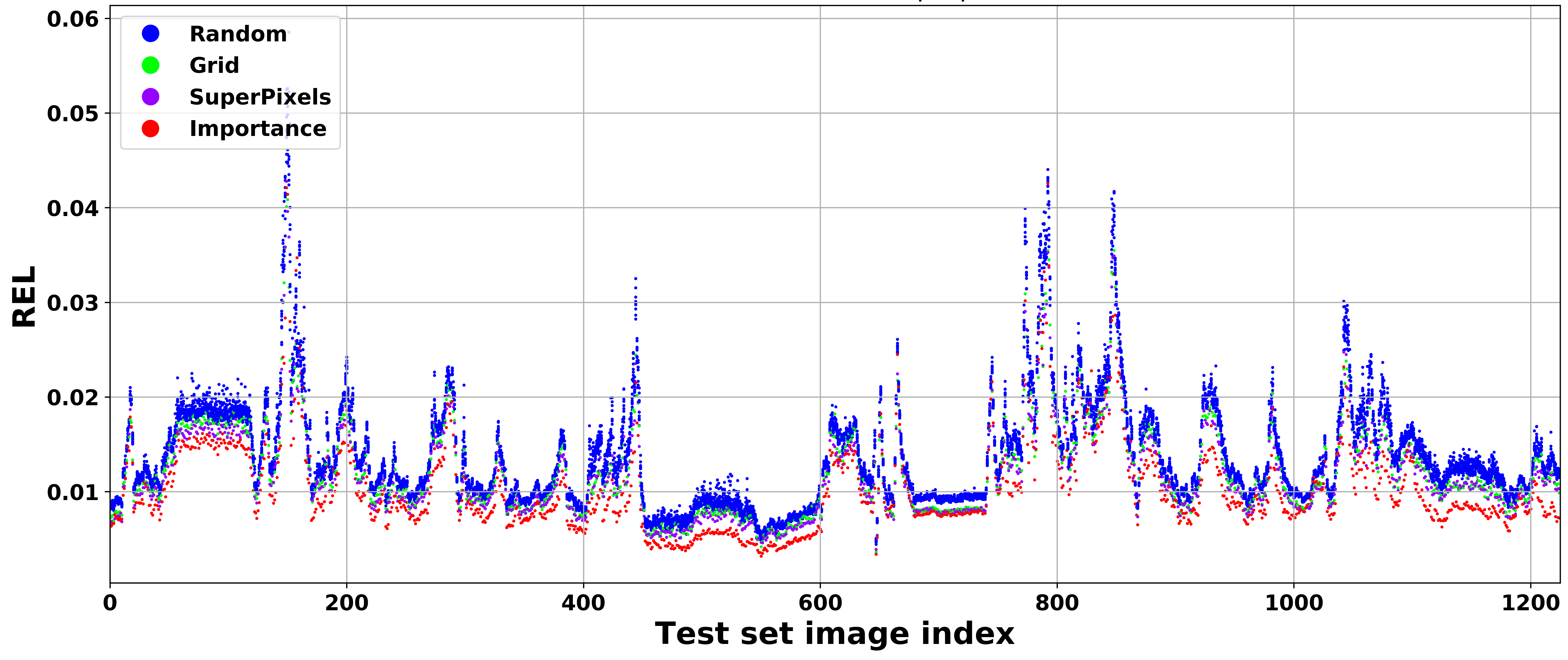}
    \caption[Reconstruction results - REL]{REL}\vspace{-0.3cm}
    \label{fig:res_rel}
  \end{subfigure}
\caption[Gaussian Sampling]{
Four different sampling patterns and the corresponding reconstruction networks were used to reconstruct depth for each of the images in the test set. Our importance-based sampling pattern (red) consistently achieves lower reconstruction error.}
\label{fig:reconstruction_results}
\end{centering}
\end{figure}

\definecolor{MyGreen}{rgb}{0,1.0,0}
\definecolor{MyPurple}{rgb}{0.5,0.0,1.0}
\begin{table}
\begin{centering}
\scalebox{0.6}{
 \begin{tabular}[width=1.0\columnwidth]{| c | c | c | c | c |}
 \hline
 Plot Color & Sampling Pattern & Reconstruction Network & RMSE [m] & REL \\
 \hline\hline
 \textcolor{blue}{Blue} & Random & $DepthReconRand$ & 1.2760 & 0.0138\\ 
 \hline
 \textcolor{MyGreen}{Green} & Grid & $DepthReconGrid$ & 1.2718 & 0.0122 \\ 
 \hline
 \textcolor{MyPurple}{Purple} & SuperPixels & $DepthReconSP$ & 1.2405 & 0.0118\\ 
 \hline
 \textcolor{red}{Red} & Importance (ours) & $DepthReconQ$ & \textbf{0.8045} & \textbf{0.0104}\\
 \hline
  - & Importance Oracle & $DepthReconQ$ & 0.4538 & 0.0082\\ 
 \hline
  - & Gofer \etal \cite{gofer2021adaptive}\protect\footnotemark & Ma \etal \cite{ma2018sparse} & (1.7558) & (0.1940) \\ 
 \hline
\end{tabular}}
\caption[Reconstruction results average]{
Average error values for each of the sampling patterns shown in \cref{fig:reconstruction_results}. Importance Oracle pattern was constructed based on the original $Q$ maps instead of the generated $\hat{Q}$ to evaluate the performance decrease due to generation defects.\vspace{-0.5cm}}
\label{tab:results}
\end{centering}
\end{table}
\interfootnotelinepenalty=10000
\footnotetext{The comparison with this method is approximate since a different depth reconstruction network (by Ma and Karaman \cite{ma2018sparse}) was used in the original implementation.}

Synthia dataset \cite{Ros_2016_CVPR} consists of 5 different driving sequences in urban environment (labeled as 1,2,4,5,6). Since in this work we do not attempt to overcome challenging visibility conditions, we chose the summer version. For each vehicle position, there are 8 RGBd couples provided for 8 different camera directions. We have used left-front and left-rear directed images of the 5 sequences, since sideways directed images do not provide scenes of a wide depth range, and the remaining front and rear directed images provide very little additional variation. Sequences 1,2,4,6 were used for training (7483 RGBd couples), while the beginning of sequence 5 was used for validation (300 RGBd couples) and the rest of it for testing (1224 RGBd couples).

Each color in the scatter plots of Fig. \ref{fig:res_rmse} and Fig. \ref{fig:res_rel} represents a sampling pattern and its corresponding reconstruction network as shown in Table \ref{tab:results}. The scatter plots shows the reconstruction error metric $M$ for each of the images in the test set. Each depth $GT$ image in the test set is sampled with 10 different random patterns, 1 grid pattern, 1 superpixels pattern, and 1 importance pattern and the depth reconstruction is done using the corresponding network. Therefore, for each point in the horizontal axis there are 13 points in the vertical axis: 10 blue, 1 green, 1 purple, and 1 red.

As shown in Fig. \ref{fig:reconstruction_results} and Table \ref{tab:results}, our adaptive sampling pattern consistently leads to a lower reconstruction error compared to other patterns. 
It achieves a reduction of 37\% in terms of RMSE and 25\% in terms of REL, compared to the average non-adaptive random pattern.
In addition, we show a considerable advantage over previous adaptive state-of-the-art method \cite{gofer2021adaptive}. This paper reports impressive results, but the proposed algorithm in its original form suffered a significant performance loss due to changes in experiment settings.

\begin{figure}
\centering
\includegraphics[width=0.95\columnwidth]{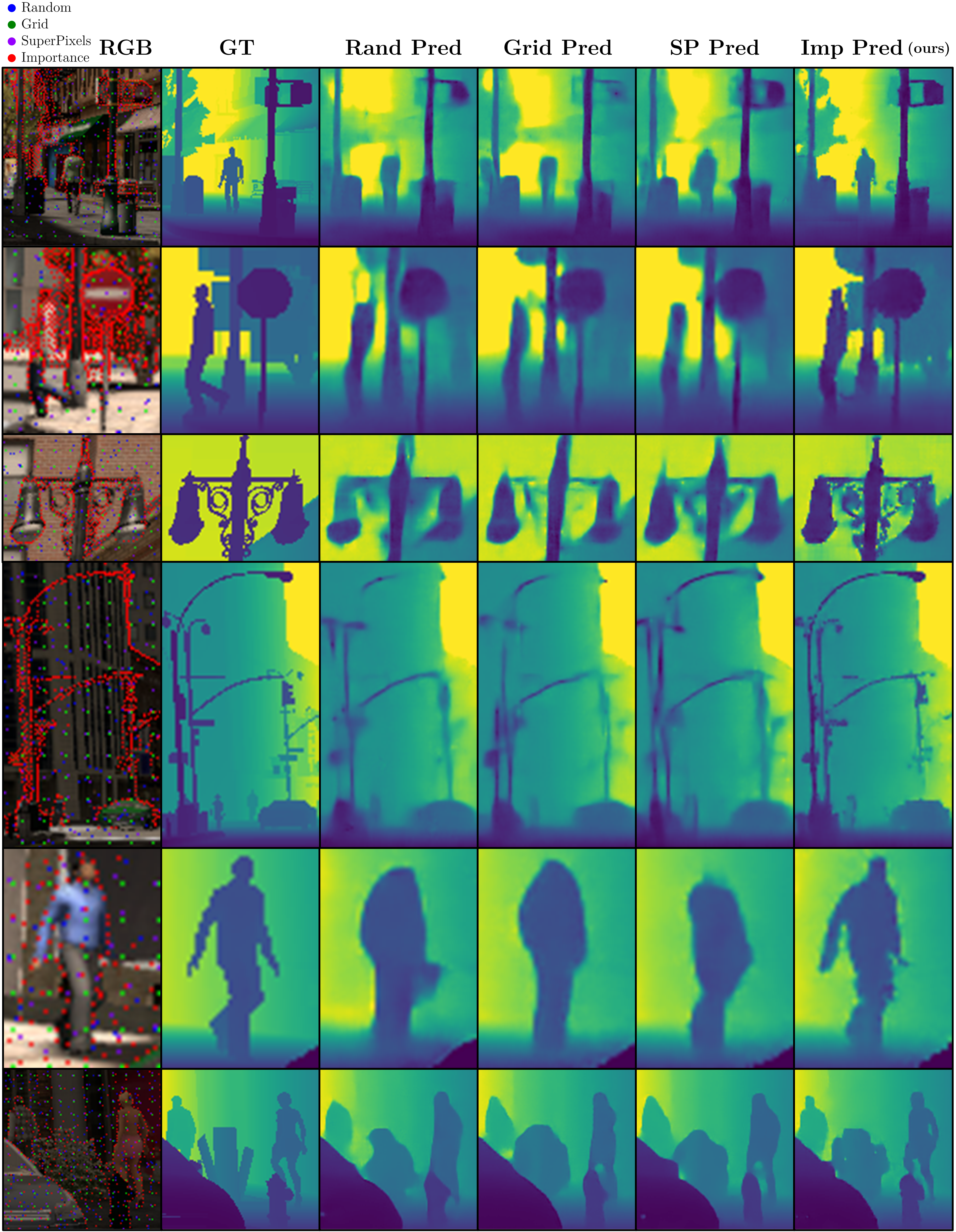}
\caption{Qualitative Results. Four different sampling patterns are marked on RGB images (crops of objects in different scenes). The corresponding depth prediction results can be compared to GT.}
\label{fig:qualitative}
\vspace{-0.2cm}
\end{figure}

Examples of qualitative results are shown in \cref{fig:qualitative}. Our importance-based sampling pattern (red) is designed to extract information which is essential for reducing a specific error metric. This patterns leads to sharper and more accurate depth prediction results (right column) than other sampling strategies.

Visually it seems that the $RGB2Q$ model generates accurate $\hat{Q}$ maps. We have tested our proposed approach for guidance of adaptive depth sampling algorithms without the effect of information degradation posed by the differences between $Q$ and $\hat{Q}$. In this test we treated the original $Q$ maps as $\hat{Q}$ (skipped step 3). Naturally, this information is not available in inference, and this was done in order to evaluate the overall concept. We denote the sampling patterns that result in this process as Oracle Importance, since they imitate a perfect generation of $\hat{Q}$ maps. As our results show (\cref{tab:results}), such sampling leads to a significantly lower reconstruction error.

\section{Conclusion}
\label{sec:conclusion}
In this work we propose a method for guidance of adaptive depth sampling algorithms. Since the sample selection problem is non-differentiable, the design of effective end-to-end architectures is challenging. Our proposed guidance technique relies on statistical estimation of the expected value of an error measure, for a specific predictor. The estimated error map of a predictor is used to guide the sampling process. We provide a method for computing the empirical error measure and for estimating it using RGB side information. A depth sampling algorithm is then suggested which is guided by our proposed map. An iterative process is used to choose the most relevant samples, with adequate spacing between them (Gaussian Sampling). A very coarse grid pattern is added to provide increased robustness.
Our results show a significant improvement over other sampling patterns, in terms of depth reconstruction accuracy. Qualitatively, we obtain an increase of the effective sampling resolution,  yielding sharper edges and finer details.
Our guidance technique offers large flexibility and can be applied for various sensors, predictors, error measures and sampling budgets. A disadvantage of this method is that it is fully guided and cannot operate without side information (such as RGB, thermal or radar data).

{\small
\bibliographystyle{ieee_fullname}
\bibliography{egbib}
}

\end{document}